\documentclass[sigconf]{acmart}
\usepackage{graphicx}
\usepackage{textcomp}
\usepackage{xcolor}
\usepackage{multirow}
\usepackage{booktabs}
\usepackage{cleveref}
\usepackage{pgfplots}
\usepackage{enumitem}
\usepackage{algorithm}
\usepackage{algorithmic}
\AtBeginDocument{%
  }

\copyrightyear{2026}
\acmYear{2026}
\setcopyright{cc}
\setcctype{by-nc-nd}
\acmConference[ICCAD '26]{IEEE/ACM International Conference on Computer-Aided Design}{November 08--12, 2026}{San Jose, CA, USA}
\acmBooktitle{IEEE/ACM International Conference on Computer-Aided Design (ICCAD '26), November 08--12, 2026, San Jose, CA, USA}
\acmDOI{10.1145/3831252.3834021}
\acmISBN{979-8-4007-2873-0/2026/11}
\begin{document}

\title{GNN-Guided Local Refinement for Efficient Layout Decomposition with DSA-MP}
\thanks{\textsuperscript{\textdagger}Corresponding author: Shisheng Xiong. This work is sponsored by the National Key Research and Development Program of China (No. 2024YFA1209000).}

\author{%
\begin{tabular}{c}
Yu Feng$^{1}$ \quad Liyi Zhou$^{1}$ \quad Haoyu Wang$^{1}$ \quad Yiming Xu$^{1}$ \quad
Hao Geng$^{2}$ \quad Shaozhi Deng$^{3}$ \\[-0.05em]
Ningsheng Xu$^{1}$ \quad Shisheng Xiong$^{1,\dagger}$ \\[0.15em]
$^{1}$Fudan University \quad $^{2}$ShanghaiTech University \quad $^{3}$Sun Yat-sen University
\end{tabular}}

% \begin{abstract}
% Multiple patterning lithography (MPL) has been widely used to improve lithography resolution in advanced nodes, but it faces fundamental limits as feature sizes continue to shrink. Directed self-assembly (DSA) offers a cost-effective complementary solution, extending the resolution of optical lithography to sub-10 nm and beyond. Integrating DSA into MPL (DSA-MP) further enhances the resolution of contact/via patterning. Nevertheless, growing design complexity exposes the limits of existing decomposition methods, which struggle to balance solution quality and runtime, and neglect edge-to-edge spacing constraints of DSA templates, compromising manufacturability. To address these issues, we introduce a new conflict graph model capturing the previously overlooked constraints and develop a novel decomposition framework powered by the graph neural network, enabling efficient and scalable layout decomposition. Experiments demonstrate that our method achieves higher solution quality with faster convergence than state-of-the-art methods, while maintaining robustness across varying numbers of vias.
% \end{abstract}
\begin{abstract}
Multiple patterning lithography (MPL) has been widely used to improve lithography resolution in advanced nodes, but it faces cost limits as feature sizes shrink. Directed self-assembly (DSA) offers a cost-effective complement, extending the resolution of optical lithography to sub-10 nm and beyond. Integrating DSA into MPL (DSA-MP) further enhances the resolution of contact/via patterning. Nevertheless, growing design complexity exposes the limits of existing decomposition methods, which struggle to balance solution quality and runtime, and neglect edge-to-edge spacing constraints of DSA templates, compromising manufacturability. To address these issues, we introduce a new conflict graph model capturing the previously overlooked constraints and develop a local refinement strategy powered by the graph neural network, enabling efficient and scalable layout decomposition. Experiments demonstrate that our method achieves higher solution quality with faster convergence than state-of-the-art methods, while maintaining robustness across layout scales, design styles, technology nodes, and varying mask counts.

\end{abstract}

\maketitle 

\fancyhead{}

\section{Introduction}
Multiple patterning lithography (MPL) has been widely adopted in advanced nodes to overcome the resolution limits of optical lithography~\cite{li2020openmpl}. By assigning closely spaced features to different masks, MPL avoids patterning conflicts. However, as feature sizes shrink, the number of required masks increases, leading to higher fabrication costs. To address this issue, directed self-assembly (DSA) has emerged as a promising and cost-effective alternative~\cite{lambrechts2018extending,wu2023lamellar}. When combined with MPL (DSA-MP), features that would otherwise conflict can share the same mask if they satisfy DSA grouping rules, effectively reducing conflicts. The underlying mechanisms and rules of DSA are detailed in \Cref{PA}.

DSA-MP holds great promise for via patterning at sub-10 nm nodes, but the associated layout decomposition problem is substantially harder than conventional MPL. Beyond assigning each via to a mask, the solver for DSA-MP must also decide which conflicting vias to group together. Grouped vias can be patterned on the same mask via DSA, effectively reducing conflicts. However, the grouping and mask assignment decisions are tightly coupled, and many valid groupings typically exist, making the search for the optimal combination far more complex. This problem is already NP-hard \cite{yu2015layout}, and the difficulty further escalates in advanced nodes, where the conflict-to-via ratio has grown from around 1:2 \cite{badr2016mask,lin2019novel,yang2017effective,kuang2016simultaneous} to 2:1 or even 3:1, reflecting much denser conflict relationships. This increase in conflict density makes prior methods difficult to scale. Exact approaches such as integer linear programming (ILP) guarantee optimality but incur prohibitive runtime. Reformulating the problem as an exact cover matrix leads to the exact cover matrix-based ILP (EILP) \cite{lin2019novel, knuth2000dancing}, which reduces problem size but remains computationally expensive. In contrast, heuristic methods like DLXplus \cite{lin2019novel}, while faster, often sacrifice solution quality. This leads to a fundamental dilemma: how can we achieve the high solution quality of ILP while avoiding its prohibitive computational cost?  

% The key to breaking out of this dilemma lies in the ILP solving process itself. While a full ILP run takes hours, the solver typically reaches a high-quality feasible solution during the early solving process. We observe that the vast majority of via assignments have already stabilized at this stage and need not be modified to minimize the overall objective. The remaining runtime is spent resolving a small number of uncertain vias. This imbalance inspires a shift in our approach. Instead of running the solver to completion, we exploit features accumulated during the early solving process to predict which vias remain uncertain. If we can identify this critical subset, the solver can re-optimize only those vias on a drastically reduced search space, saving enormous computational effort.

% The key to breaking out of this dilemma lies in the ILP solving process itself. While a full ILP run takes hours, the solver typically reaches a high-quality feasible solution and generates rich intermediate features during the early solving process. We observe that reassigning the masks of only a small fraction of vias in this early solution is sufficient to reach the optimal objective value. If this critical subset can be identified, the solver need only re-optimize those vias on a drastically reduced search space, saving enormous computational effort.  We therefore use the ILP solver not as an end-to-end optimizer, but as a source of early-stage features that help predict which vias require mask reassignment to reach the optimum.
\begin{figure}[!t]
\centering
\includegraphics[width=1.0\linewidth]{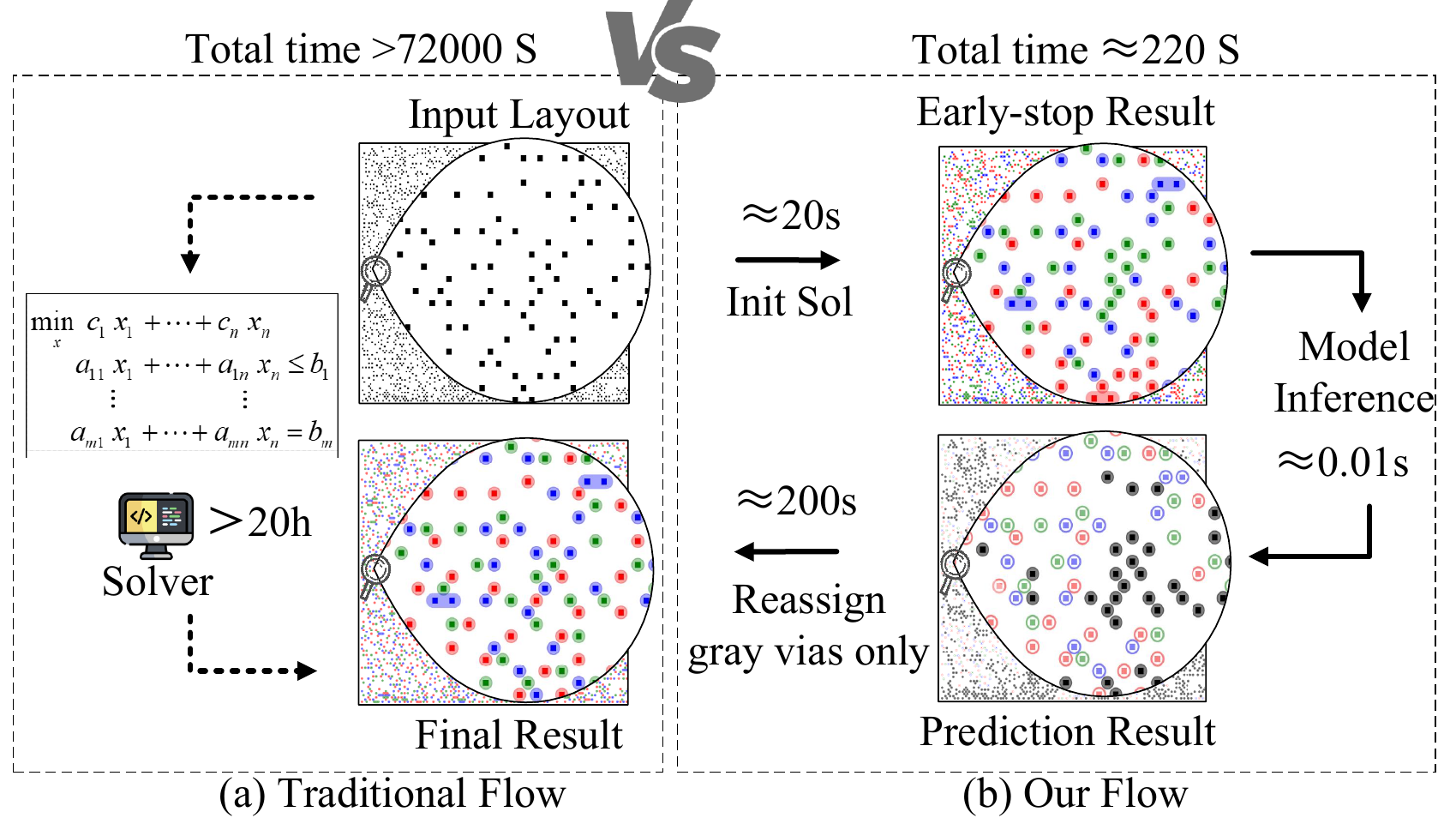}
% \caption{(a) Traditional flow: direct solving from scratch until optimal or timeout. (b) Instead of solving large-scale problems directly, makes full use of intermediate features produced by a time-limited solver to shrink the problem size. }
\vspace{-5mm} % 调整这个值
\caption{(a) Traditional flow: solve from scratch until optimal or timeout. (b) Leveraging features from the initial solving phase, our flow predicts vias for mask reassignment. The solver reassigns masks only for these vias, keeping all others fixed, significantly speeding up convergence.} 
% \caption{(a) The flow of our proposed local refinement framework. (b) The layout is first modeled as a conflict graph and then formulated as an ILP. A time-limited solver produces (c) an initial solution with solver-derived features, which guide the prediction of (d) vias requiring mask reassignment. Only these vias are re-optimized while others remain fixed to obtain (e) the final result.} 
% Workflow of the proposed ForgePattern. (white regions represent the metal shapes)
\vspace{-3mm} % 调整这个值
\label{introduction1}
\end{figure}

The key to breaking out of this dilemma lies in the ILP solving process itself. While a full ILP run takes hours, the solver typically reaches a high-quality feasible solution and accumulates rich intermediate solving records during the early stage of the ILP solve. We observe that reassigning only a small fraction of vias in this early solution is sufficient to match the optimal objective value. If we can identify this critical subset, the solver can re-optimize only those vias on a drastically reduced search space, saving enormous computational effort. This insight calls for a shift in perspective: instead of treating the ILP solver merely as an end-to-end optimizer that runs for hours to deliver a final answer, we exploit the features accumulated during its early solving phase to predict which vias require mask reassignment to reach the optimum.

\begin{figure}[!t]
\centering
\includegraphics[width=0.80\linewidth,]{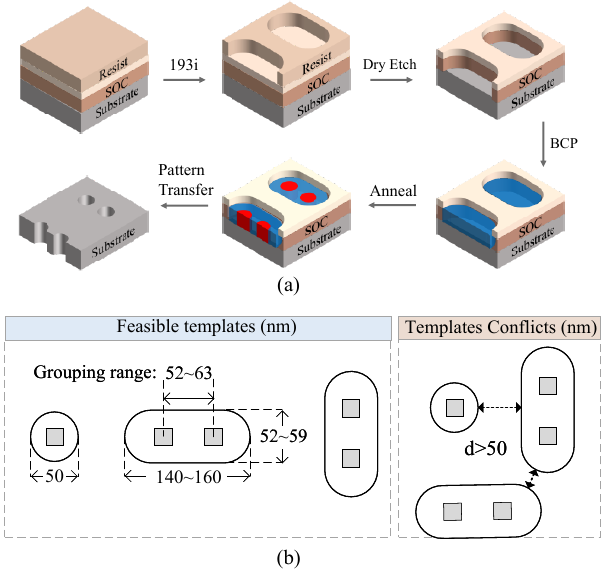}
% \caption{(a) DSA process flow: 193i lithography and etching define guiding templates on the substrate. A block copolymer (e.g., PS-b-PMMA) is deposited and thermally annealed to induce self‑assembly within the templates, and the resulting patterns are transferred into the underlying layers. (b) Feasible templates: adjusting template size enables different via pitches, but the grouping range is limited by the natural pitch $L_0$ of the BCP. Moreover, a minimum edge-to-edge spacing between templates is required to ensure stable patterning. }
\vspace{-4mm} 
\caption{(a) DSA process flow. (b) Feasible templates: by adjusting template size, vias with different pitches can be fabricated, but the grouping range is limited by the natural pitch $L_0$ of the BCP. Moreover, a minimum edge-to-edge spacing between templates is required to ensure stable patterning. } 
% Workflow of the proposed ForgePattern. (white regions represent the metal shapes)
\vspace{-3mm} 
\label{process}
\end{figure}

This idea drives our approach. As illustrated in Figure 1, a time-limited ILP solver first provides an initial solution and intermediate features. A graph neural network (GNN) \cite{kipf2016semi} trained on these features then predicts which vias are likely to benefit from reassignment. The solver re-optimizes only the selected vias while keeping all others fixed, converging to high-quality solutions in a fraction of the original runtime. Analogous to a lever, the solver's intermediate features serve as a fulcrum: a small, targeted effort produces a disproportionately large improvement. However, efficient solving alone is not enough if the underlying conflict model is inaccurate. Existing methods \cite{badr2016mask,lin2019novel,yang2017effective,kuang2016simultaneous} define conflicts based on center-to-center distances between vias, a metric carried over from conventional MPL.  In DSA-MP, 193i lithography fabricates guiding templates instead of direct vias, and these templates may take non-circular shapes \cite{ou2016concurrent}. Simply evaluating manufacturability by checking center-to-center distances between vias fails to reflect the true process constraints. Since the final via quality is highly dependent on template geometry, sufficient spacing between templates is required to provide a correction margin against proximity effects and ensure stable template formation. Therefore, we construct a conflict graph that explicitly captures edge-to-edge template spacing and refine the EILP formulation accordingly.
Our major contributions are as follows:
\begin{itemize}
    \item We propose a novel conflict graph model for DSA-MP and refine the formulation of EILP to account for edge-to-edge template conflicts.
    \item We propose a GNN-guided local refinement strategy, where the trained GNN model efficiently selects vias for refinement while keeping others fixed, significantly accelerating convergence.
    
    % \item We propose generating a diverse layout library that complies with advanced-node design rules to train the GNN model, thereby enhancing its accuracy and generalization.

    \item Our method outperforms state-of-the-art works by achieving about 45$\times$ speedup with only 0.1\% cost increase on 7~nm synthetic benchmarks, while generalizing well to other 7~nm designs and 5~nm industrial layouts.
\end{itemize}
% Unlike prior works that only consider center-to-center violations, our formulation uniformly treats all conflicts as template edge spacing constraints.
% To overcome the scarcity of real layout data, we further generate synthetic layouts following the same design rules to provide additional training data for the GNN.
\section{Preliminaries}
\label{pre}
% \begin{figure}[!t]
% \centering
% \includegraphics[width=0.88\linewidth,]{MPL/figs/dsa_process_intro_V3.pdf}
% % \caption{(a) DSA process flow: 193i lithography and etching define guiding templates on the substrate. A block copolymer (e.g., PS-b-PMMA) is deposited and thermally annealed to induce self‑assembly within the templates, and the resulting patterns are transferred into the underlying layers. (b) Feasible templates: adjusting template size enables different via pitches, but the grouping range is limited by the natural pitch $L_0$ of the BCP. Moreover, a minimum edge-to-edge spacing between templates is required to ensure stable patterning. }
% \caption{(a) DSA process flow. (b) Feasible templates: by adjusting template size, vias with different pitches can be fabricated, but the grouping range is limited by the natural pitch $L_0$ of the BCP. Moreover, a minimum edge-to-edge spacing between templates is required to ensure stable patterning. } 
% % Workflow of the proposed ForgePattern. (white regions represent the metal shapes)
% \label{process}
% \end{figure}

\subsection{Brief Introduction to DSA}
\label{PA}
Directed self-assembly (DSA) utilizes block copolymers (BCPs) guided by lithographically defined templates for sub-lithographic patterning, as shown in \Cref{process}(a). The process involves template definition, BCP deposition, annealing, and pattern transfer, enabling the formation of dense and uniform vias~\cite{bekaert2015n7}.

\begin{figure}[!t]
\centering
\includegraphics[width=0.95\linewidth]{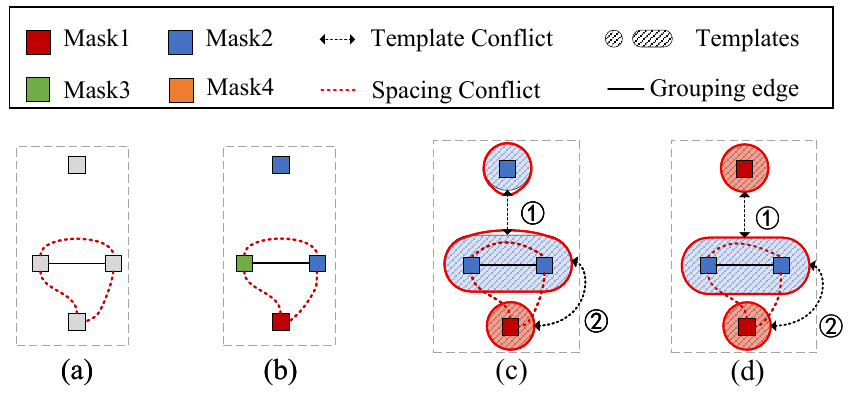}
\caption{(a) Original layout. (b) Decomposition with triple patterning. (c) Decomposition with DSA-MP (The red solid line shows the lithography simulation result from S-litho). (d) Another decomposition with DSA-MP.} 
% Workflow of the proposed ForgePattern. (white regions represent the metal shapes)
\label{Conflict Example}

\end{figure}

As illustrated in \Cref{process}(b), we restrict feasible configurations to 1-D templates (n × 1 holes, $n \leq 2$) to ensure high yield and robustness, since templates with fewer holes generally provide more stable and reliable patterning \cite{lin2019novel}. The spacing of vias/contacts formed by the DSA process within a template is governed by the natural pitch $L_0$ of the BCP, which limits the grouping distance to a specific range ($52\text{--}63$ nm in this work). In addition, sufficient edge-to-edge spacing between adjacent templates must also be reserved. Without sufficient spacing, OPC and other correction techniques lack the room to counteract proximity effects, making it impossible to ensure stable and accurate template profiles. 
However, the edge-to-edge spacing constraints of templates were often neglected in previous studies. For example, in \Cref{Conflict Example}(c) and \Cref{Conflict Example}(d), previous work considered both assignments feasible, 
but in reality only \Cref{Conflict Example}(d) satisfies manufacturability, as the simulated contour in \Cref{Conflict Example}(c) fails to match the required template profile. The contours are obtained using the rigorous lithography simulator S-Litho~\cite{synopsys_slitho}.
% However, these constraints were often oversimplified in previous studies, motivating the need for a new conflict graph model that explicitly captures them (see Section II-B).

% \begin{figure}[!t]
% \centering
% \includegraphics[width=0.95\linewidth]{MPL/figs/conflctgraphV12.pdf}
% \caption{(a) Original layout. (b) Decomposition with triple patterning. (c) Decomposition with DSA-MP. (d) Another decomposition with DSA-MP.} 
% % Workflow of the proposed ForgePattern. (white regions represent the metal shapes)
% \label{Conflict Example}
% \end{figure}

\begin{figure}[!t]
\setlength{\abovecaptionskip}{3pt}
\setlength{\belowcaptionskip}{3pt}
\centering
\includegraphics[width=0.85\linewidth]{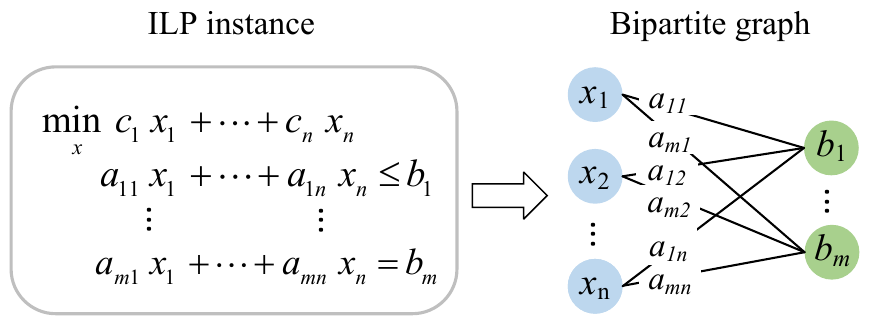}
\caption{Transforming an ILP instance to a bipartite graph. 
 } 
% Workflow of the proposed ForgePattern. (white regions represent the metal shapes)
\label{bipartite}
\end{figure}

\subsection{Problem Formulation}
\textbf{Definition 1.} (Groupable vias): Two vias are defined as groupable if their center-to-center distance falls within the feasible DSA grouping range and they are aligned either horizontally or vertically. Such vias can be manufactured together by a two-hole guiding template.
% {Definition 2.} (Spacing conflicts): Lithography spacing conflicts and edge-to-edge template spacing conflicts can be unified as edge-to-edge template spacing conflicts.

% \textbf{Problem 1.} (Guiding template
% and mask assignment (TMA)): Given a via layout, a minimum edge-to-edge template spacing $d_{min}$, $K$ available masks, and a set of feasible guiding templates with template costs, the problem is to assign each via to a template and each template to one of the $K$ masks. If the minimum edge-to-edge distance between two templates on the same mask is smaller than $d_{min}$, a template spacing conflict occurs. The objective is to minimize the total number of template spacing conflicts and the total template cost.
\textbf{Problem 1.} (Guiding template and mask assignment (TMA)):
Given a via layout with $|V|$ vias, a minimum edge-to-edge template spacing $d_{\min}$, $K$ available masks, and a set of feasible guiding templates with associated template costs, a template spacing conflict occurs if the edge-to-edge distance between two templates on the same mask is smaller than $d_{\min}$. The problem is to assign each via to a guiding template and each template to one of the $K$ masks, such that the total number of template spacing conflicts and the total template cost are minimized.

\subsection{Bipartite Graph Representation}
% \label{bipartite}
The bipartite graph representation of integer linear programs (ILPs) was proposed by \cite{gasse2019exact}, which losslessly transforms the ILP instance into a bipartite graph (see \Cref{bipartite}).
 In this graph, the left part contains $n$ \emph{variable nodes} $X = \{x_1, \dots, x_n\}$ corresponding to the decision variables, and the right part contains $m$ \emph{constraint nodes} $B = \{b_1, \dots, b_m\}$ corresponding to the linear constraints. An edge $(i,j) \in E$ between variable node $i$ and constraint node $j$ exists whenever the $i$-th variable appears in the $j$-th constraint, with the edge weight storing the coefficient $a_{ij}$. This representation preserves the original problem structure and provides a natural input for GNN models.

\section{Proposed Graph Model and ILP Formulation}
\label{sec:graph_model}
\begin{figure}[!t]
\centering
\includegraphics[width=0.9\linewidth]{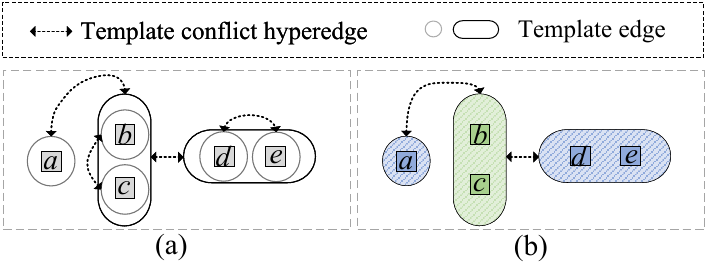}
\caption{(a) Our graph model. It unifies lithography spacing and template-related conflicts as edge-to-edge constraints between templates. Each via can be manufactured by a single-hole template, while vias connected by a two-hole template edge may alternatively use the two-hole template. Template conflict hyperedges represent all possible template conflicts. (b) An example of a feasible mask and template assignment.} 
% Workflow of the proposed ForgePattern. (white regions represent the metal shapes)
\label{hybrid graph}
\vspace{-4mm}
\end{figure}
% \begin{figure}[!t]
% \centering
% \includegraphics[width=0.9\linewidth]{MPL/figs/Exact coverV5.pdf}
% \caption{Exact cover matrix for TMA of DSA-DP. The listed entries indicate 1’s entries. The purple dashed lines highlight the selected rows corresponding to the TMA solution shown in \Cref{hybrid graph}(b).} 
% % Workflow of the proposed ForgePattern. (white regions represent the metal shapes)
% \label{exact cover}
% % \vspace{-5pt}
% \end{figure}
% \textbf{Definition 2.} (Hybrid grouping/conflict graph.): A hybrid grouping/conflict graph represents a via layout in DSA-MP hybrid lithography, capturing both spacing and template constraints. It contains three types of edges: spacing edges for lithography conflicts between vias, template conflict edges for conflicts between templates, and template hyperedges linking vias that can be manufactured together by a feasible multi-via guiding template.

To better capture lithography spacing and template-related conflicts in a given layout, we unify them as edge-to-edge constraints between templates and introduce a hybrid grouping/conflict graph, as shown in \Cref{hybrid graph}. In this graph, each vertex represents a via, template edges indicate that a via can be fabricated by a single-hole template alone or paired with another via to form a feasible two-hole template, and template conflict hyperedges denote cases where the edge-to-edge distance between templates is less than $d_{min}$. Built on this compact graph model, we develop a novel and concise ILP formulation that explicitly accounts for edge-to-edge template spacing constraints to optimally solve the TMA problem. The notations used in our ILP formulation are
listed as follows.

\begin{figure}[!t]
\centering
\includegraphics[width=0.9\linewidth]{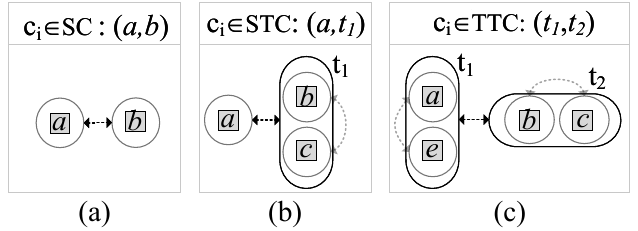}
\caption{Illustration of three types of edge-to-edge template spacing conflicts, each with a representative $c_i$ highlighted by the black dashed arrow. (a) $SC$: taking $c_i = (a, b)$ as an example, a conflict between two single-hole templates. (b) $STC$: taking $c_i = (a, t_1)$ as an example, a conflict between single-hole template $a$ and two-hole template $t_1$. (c) $TTC$: taking $c_i = (t_1, t_2)$ as an example, a conflict between two-hole templates $t_1$ and $t_2$.}
% \caption{Illustration of three types of edge-to-edge template spacing conflicts, each with a representative $c_i$ highlighted by the black dashed arrow. (a) $c_i = (a, b)$ is a representative conflict in $SC$, where two single-hole templates are in conflict. (b) $c_i = (a, t_1)$ is a representative conflict in $STC$, where a single-hole template $a$ and a double-hole template $t_1$ are in conflict. (c) $c_i = (t_1, t_2)$ is a representative conflict in $TTC$, where two double-hole templates are in conflict.}
\label{exact cover}
% \vspace{-5pt}
\vspace{-2mm}
\end{figure}

\begin{enumerate}[label=\arabic*)] %\arabic
    \item $SC$, $STC$, $TTC$: sets of edge-to-edge spacing conflict candidates, as illustrated in \Cref{exact cover}. Specifically, $SC$ represents conflicts between two single-hole templates,
    $STC$ represents conflicts between a single-hole and a two-hole template, and
    $TTC$ represents conflicts where both templates are two-hole.
  \item $\mathcal{I}_{SC},\mathcal{I}_{STC},\mathcal{I}_{TTC}$: index sets corresponding to $SC,STC,$ and $TTC$.
  \item $c_i$: Conflict $c_i$, $i \in \mathcal{I}_{SC}\cup\mathcal{I}_{STC}\cup\mathcal{I}_{TTC}$.
  \item $v_i$: The $i$-th via. 
  \item $t_j$: Template $t_j$, with two vias in a template edge.
  \item $c_i^k$: 0/1 Boolean variable. $c_i^k = 1$ if the two templates involved in conflict $c_i$ conflict at mask $k$.
  \item $v_i^k$: 0/1 Boolean variable. $v_i^k = 1$ if via $v_i$ is assigned to a single-hole template on mask $k$.
  \item $t_j^k$: 0/1 Boolean variable. $t_j^k = 1$ if a two-hole template $t_j$ is chosen, and is assigned to mask $k$.
  \item $w_c$: Conflict cost. $w_{t}$: Two-hole template cost.
\end{enumerate}

The ILP formulation is given as follows:
% \begin{align}
% \min \quad 
% & \sum_{i \in \mathcal{I}_{SC} \cup \mathcal{I}_{STC} \cup \mathcal{I}_{TTC}}
%   \sum_{k=1}^{K} w_c \cdot c_i^k
%   + \sum_{j,k} w_t \cdot t_j^k
%   \label{eq:obj} \\[6pt]
% %
% \text{s.t.} \quad 
% & \sum_{j: v_i \in t_j} \sum_{k=1}^{K} t_j^k 
%   + \sum_{k=1}^{K} v_i^k = 1,
%   \quad \forall i = 1, \ldots, |V|
%   \label{eq:via} \\[6pt]
% %
% & \sum_{j: (u,v) \in c_i \wedge u \in t_j, v \notin t_j} t_j^k
%   + \sum_{j: v_j \in c_i} v_j^k \leq 1 + c_i^k,\\
% &  \hspace{2.5cm} \forall i \in \mathcal{I}_{SC},\ \forall k = 1, \ldots, K
%   \label{eq:conflict_SC} \nonumber \\[6pt]
% %
% & \sum_{j: v \in c_i \wedge v \in t_j} t_j^k
%   + \sum_{j: v_j \in c_i} v_j^k
%   + \sum_{j: t_j \in c_i} t_j^k \leq 1 + c_i^k,\\
% &  \hspace{2.5cm} \forall i \in \mathcal{I}_{STC},\ \forall k = 1, \ldots, K  \nonumber
% \\[6pt]
% %
% & \sum_{j: t_j \in c_i} t_j^k \leq 1 + c_i^k,  \forall i \in \mathcal{I}_{TTC},\ \forall k = 1, \ldots, K \label{eq:conflict_TTC}
% \end{align}
\begin{align}
\min \quad 
& \sum_{i \in \mathcal{I}_{SC} \cup \mathcal{I}_{STC} \cup \mathcal{I}_{TTC}}
  \sum_{k=1}^{K} w_c \cdot c_i^k
  + \sum_{j,k} w_t \cdot t_j^k
  \label{eq:obj} \\[6pt]
\text{s.t.} \quad 
& \sum_{j: v_i \in t_j} \sum_{k=1}^{K} t_j^k 
  + \sum_{k=1}^{K} v_i^k = 1,
  \quad \forall i = 1, \ldots, |V|
  \label{eq:via} \\[6pt]
& \sum_{j: v_j \in c_i} v_j^k \leq 1 + c_i^k, \forall i \in \mathcal{I}_{SC},\ \forall k = 1, \ldots, K
  \label{eq:conflict_SC}  \\[6pt]
& \sum_{j: v_j \in c_i} v_j^k
  + \sum_{j: t_j \in c_i} t_j^k \leq 1 + c_i^k, \nonumber \\
&  \hspace{2.9cm} \forall i \in \mathcal{I}_{STC},\ \forall k = 1, \ldots, K  
  \label{eq:conflict_STC} \\[6pt]
& \sum_{j: t_j \in c_i} t_j^k \leq 1 + c_i^k,  \forall i \in \mathcal{I}_{TTC},\ \forall k = 1, \ldots, K \label{eq:conflict_TTC}
\end{align}

\begin{figure*}[!t]
\centering
\setlength{\abovecaptionskip}{3pt}
\setlength{\belowcaptionskip}{3pt}
\includegraphics[width=0.89\linewidth]{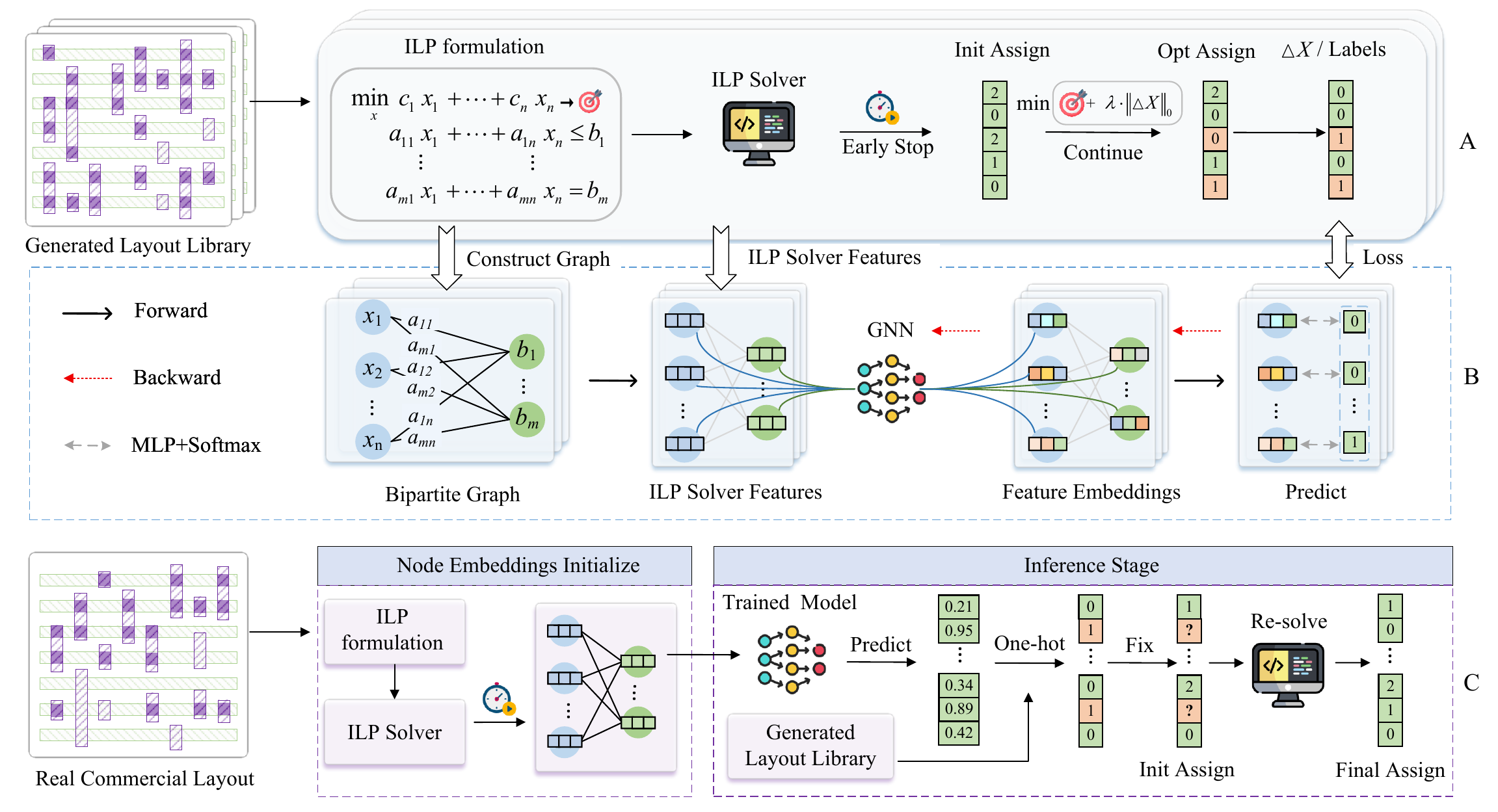}
\caption{Our proposed framework. (A)  Synthetic layouts are generated based on design rules extracted from the ASAP7 PDK. A time-limited ILP solver produces an initial solution, and a constraint on the count of via changes is added to identify the minimal set of vias requiring reassignment, forming the Vias-to-Refine dataset.
(B) The ILP formulation is represented as a bipartite graph, and solver-derived features initialize a GNN trained on the dataset from (A) to predict vias requiring reassignment.
(C) For a test layout, a time-limited ILP solver provides an initial solution. Based on GNN-predicted probabilities, a subset of vias is sampled for reassignment, while others are fixed, and only the selected vias are re-optimized.
}
% Workflow of the proposed ForgePattern. (white regions represent the metal shapes)
\label{method}
\vspace{-1mm}
\end{figure*}

\textbf{Objective \eqref{eq:obj}.} 
The objective minimizes the number of unresolved conflicts and the number of selected two-hole templates. To give priority to conflict minimization, \(w_c\) is set much larger than 
\(w_t\).

\textbf{Constraint \eqref{eq:via}.} 
Each via $v_i$ must be covered by exactly one template on exactly one mask: 
either assigned to a single-hole template on some mask $k$ (i.e., $v_i^k = 1$), 
or covered by a two-hole template $t_j$ assigned to mask $k$ (i.e., $t_j^k = 1$), 
with the two cases being mutually exclusive.

\textbf{Constraint \eqref{eq:conflict_SC}.} 
For each $SC$ conflict candidate $c_i$ with $i \in \mathcal{I}_{SC}$ and each mask $k$,
the summation $\sum_{j: v_j \in c_i} v_j^k$ runs over the two vias involved in $c_i$, 
each fabricated by a single-hole template. 
If both are assigned to mask $k$, the sum equals~2, forcing $c_i^k = 1$ 
and incurring a conflict penalty in the objective. 
If they are assigned to different masks, the sum is at most~1 
and $c_i^k$ can remain~0.

\textbf{Constraint \eqref{eq:conflict_STC}.} 
For each $STC$ conflict candidate $c_i$ with $i \in \mathcal{I}_{STC}$ and each mask $k$, 
the constraint jointly considers both template types involved in $c_i$: 
$\sum_{j: v_j \in c_i} v_j^k$ accounts for the single-hole template, 
and $\sum_{j: t_j \in c_i} t_j^k$ accounts for the two-hole template. 
If both templates are assigned to the same mask $k$, 
the left-hand side equals~2, forcing $c_i^k = 1$. 
Otherwise, $c_i^k$ can remain~0.

\textbf{Constraint \eqref{eq:conflict_TTC}.} 
For each $TTC$ conflict candidate $c_i$ with $i \in \mathcal{I}_{TTC}$ and each mask $k$, 
$c_i$ involves two templates, both of which are two-hole templates. 
The summation $\sum_{j: t_j \in c_i} t_j^k$ counts how many of them are assigned to mask $k$. 
If both are selected and assigned to the same mask, 
the sum equals~2, forcing $c_i^k = 1$ and incurring a conflict penalty. 
Otherwise, $c_i^k$ can remain~0.

\section{Algorithms}
\label{alg}

\subsection{Overview}\label{AA}
To accelerate the TMA problem, we propose a GNN-based local refinement framework. As illustrated in \Cref{method}, our framework is divided into three main components. First, we construct the Vias-to-Refine Dataset (\Cref{method}A), which labels the vias that need reassignment compared to the initial solution to achieve optimal results. 
Second, we train a GNN model using the dataset (\Cref{method}B), encoding ILP-derived features into a graph to learn which vias need reassignment. Third, during inference (\Cref{method}C), vias requiring reassignment are sampled according to the probabilities predicted by the GNN, and only the variables associated with the sampled vias are re-optimized, yielding a high-quality solution with significantly reduced runtime compared to solving the full ILP from scratch.
\vspace{-3pt}
\subsection{Vias-to-Refine Dataset Preparation}
\label{BB}
Training a GNN model to predict vias requiring reassignment demands layouts with sufficient diversity in conflict graph topologies to ensure good generalization, but most real-world designs are proprietary and not publicly available. Following~\cite{yang2019automatic}, we generate synthetic layouts track-by-track based on design rules extracted from the open-source ASAP7 process design kit (PDK) ~\cite{clark2016asap7}, including wire width $w$ and track pitch $p$. Candidate vias are placed at intersections of adjacent metal layers, with enclosure and pitch constraints enforced via an assist wire layer. Wire lengths, spacings, and via placements are randomized within design-rule constraints, resulting in layouts with diverse yet design-rule-compliant conflict graph topologies. This produces a diverse set of layouts $\mathcal{L} = \{L_1, \dots, L_N\}$ covering a wide range of local conflict structures for GNN training.

% \begin{table}[htbp]
% \centering
% \caption{Description of the ILP-solve features in the bipartite graph $G = (X \cup B, E)$}
% \begin{tabular}{c|c|c}
% \hline
% \textbf{Tensor} & \textbf{Features} & \textbf{Description} \\
% \hline
% \multirow{4}{*}{$H_X$} & coef           & Objective coefficient \\ 
%                         & sol\_val       & Solution value \\ 
%                         & sol\_pool\_freq & Frequency of adding to the solution pool \\ 
%                         & point\_num     & Index of corresponding via \\
% \hline
% $H_B$ & bias  & Bias Value \\
% \hline
% $H_E$ & coef  & Constraint coefficient \\
% \hline
% \end{tabular}
% \label{tab1}
% \end{table}

For each layout $L_i$, the ILP formulation from \Cref{sec:graph_model} is solved with a time-limited early stop to obtain the initial ILP-solve features. Exploiting the inherent bipartite structure of the ILP, each layout is modeled as a bipartite graph $G = (X \cup B, E)$ as introduced in \Cref{bipartite}. The feature sets of $G$ are grouped as follows:
\begin{itemize}
  \item \textbf{Variable features $H_X$:} feature vectors for decision variables (left side of the bipartite graph in \Cref{bipartite}), including  
  (1) objective coefficient of the variable (\texttt{coef}); 
  (2) frequency with which the variable appears in the solution pool (\texttt{sol\_pool\_freq}); 
  and (3) index of the via 
  (\texttt{point num}).
  
  \item \textbf{Constraint features $H_B$:} feature vectors for the constraints (right side of the bipartite graph), including 
  (1) bias value (\texttt{bias}), normalized with respect to the constraint coefficients.
  
  \item \textbf{Edge features $H_E$:} feature vectors for edges, including 
  (1) coefficient of the decision variable in the corresponding constraint (\texttt{cons\_coef}).
\end{itemize}
All features are normalized. These node and edge features are encoded into the bipartite graph $G$ and form the input to the dataset.

% To generate labels for the dataset, the ILP is augmented with a penalty term that counts the number of vias requiring mask reassignment:
% \begin{equation}
% \min \sum_{i,k} w_c\, c_{ik} + \sum_{j,k} w_{t,j}\, t_k^j + \lambda \sum_{v \in R} r_v,
% \end{equation}
% where $r_v = 1$ indicates that via $v$ must be reassigned, and $\lambda$ is a small penalty weight (empirically set to $0.01$) to preserve the optimality of the original cost. 
% Solving this ILP yields the optimal solution $x^*$ and the minimal set of vias $\mathcal{R}^* \subseteq V$ that require reassignment.
To generate labels for the dataset, the ILP formulation in \Cref{sec:graph_model} is modified to include a penalty term that counts the number of vias whose mask assignment changes from the initial solution:
\begin{equation}
\min \sum_{i,k} w_c \cdot c_i^k + \sum_{j,k} w_t \cdot t_j^k 
+ \sum_{i=1}^{|V|} \lambda \cdot r_i,
\end{equation}
where $\lambda$ is a small penalty weight (empirically set to $10^{-4}$) to preserve the optimality of the original cost.
We introduce a binary variable $r_i\in\{0,1\}$ for via $v_i$ to capture whether its assignment changes between the initial and current solutions:
\begin{equation}
r_i = 1 - \sum_{k=1}^{K} \hat v_i^k \cdot v_i^k, \quad i=1,\dots,|V|,
\end{equation}
with $v_i^k\in\{0,1\}$ indicating whether via $v_i$ is assigned to mask $k$ in the current solution and 
$\hat v_i^k\in\{0,1\}$ denoting its assignment in the initial solution, which is a known constant.
Solving this ILP yields the optimal solution and the minimal set of vias 
$\mathcal{R}^*=\{\,v_i\mid r_i=1\,\}$ that require reassignment.

The resulting dataset consists of a bipartite graph $G$ initialized with ILP-derived features together with node-level labels. 
Each via variable $v_i^k$ in $G$ is assigned the label \texttt{True} if reassignment of $v_i$ is required and \texttt{False} if no reassignment is needed. 
All remaining variables are assigned the label \texttt{None}, indicating that they are excluded from the computation of the training loss.

% \begin{figure}[h]
% \setlength{\abovecaptionskip}{3pt}
% \setlength{\belowcaptionskip}{3pt}
% \centering
% \includegraphics[width=0.85\linewidth]{MPL/figs/layoutmpl_bipart.pdf}
% \caption{Transforming an ILP instance to a bipartite graph. 
%  } 
% % Workflow of the proposed ForgePattern. (white regions represent the metal shapes)
% \label{bipartite_v2}
% \end{figure}

\subsection{GNN Training for Via Selection}

Based on the Vias-to-Refine Dataset described in \Cref{BB}, we train a GNN model to predict which vias need reassignment. 
The model takes as input three types of tensors: variable features $H_X \in \mathbb{R}^{n \times d}$, constraint features $H_B \in \mathbb{R}^{m \times c}$, and edge features $H_E \in \mathbb{R}^{m \times n \times e}$. Here $d$ corresponds to the number of ILP-derived features for each variable (objective coefficient, solution-pool frequency, and via index, giving $d=3$ in our implementation), 
$c$ denotes the number of features per constraint (bias value, hence $c=1$), 
and $e$ represents the number of features per edge (constraint coefficient, hence $e=1$). 
All features are normalized and then projected into a latent space of dimension $64$.

\sloppy
Following the input projection, message passing is performed over the bipartite graph in the form of two interleaved half-convolutions\cite{gasse2019exact}, which alternately propagate messages between the variable and constraint nodes. 
Due to the inherent bipartite structure, our graph convolution can be decomposed into two successive passes: one propagating messages from variable nodes to constraint nodes, and the other from constraint nodes back to variable nodes. 
Formally, let $V_X$ represent the set of $n$ variable nodes and $V_B$ the set of $m$ constraint nodes. The embeddings of variable node $x_i \in V_X$ and constraint node $b_j \in V_B$ at layer $k$ are denoted by $h_{x_i}^k$ and $h_{b_j}^k$, respectively.
The update rules are defined as:
\[
h_{b_j}^{k} =
f_B^{k}\!\Big(
h_{b_j}^{k-1},
\sum_{(x_i, b_j) \in \mathcal{E}}
g_B^{k}\!\big(h_{x_i}^{k-1}, h_{b_j}^{k-1}, e_{ij}\big)
\Big),
\quad b_j \in V_B,
\]
\[
h_{x_i}^{k} =
f_X^{k}\!\Big(
h_{x_i}^{k-1},
\sum_{(x_i, b_j) \in \mathcal{E}}
g_X^{k}\!\big(h_{x_i}^{k-1}, h_{b_j}^{k}, e_{ij}\big)
\Big),
\quad x_i \in V_X,
\]
where $f_X^{k}$, $f_B^{k}$, $g_X^{k}$, and $g_B^{k}$ are implemented as 2-layer perceptrons with ReLU activations, and $e_{ij} \in \mathbb{R}^{1\times64}$ denotes the  feature tensor of the edge connecting $x_i$ and $b_j$.

After the half-convolutions, the variable node embeddings are passed through a 2-layer perceptron followed by a sigmoid activation, which outputs a probability $p_i^k \in [0,1]$ for each via variable $v_i^k$, indicating whether it requires reassignment. 

% The model is trained using binary cross-entropy loss, 
% where positive labels correspond to vias identified as needing reassignment in the dataset. 
The model is trained using a binary cross-entropy loss $\mathcal{L} = \mathrm{BCE}(y_i^k, p_i^k),$ computed over all via variables $v_i^k$,
where $y_i^k \in \{0,1\}$ denotes the ground-truth label indicating 
whether $v_i$ on mask $k$ requires reassignment. During inference, only vias predicted as positive are refined, 
thereby significantly reducing computation compared to solving the full ILP.

% \begin{algorithm}[!t]
% \caption{GNN-Based Local Refinement Inference}
% \label{alg:inference}
% \begin{algorithmic}[1]
% \REQUIRE Layout $l$, trained GNN model, temperature $\tau$, $\varepsilon$
% \ENSURE Final TMA solution

% \STATE Solve ILP with time limit to obtain initial solution $\{\hat{v}_i^k\}$ using Eq.~\eqref{eq:obj}--\eqref{eq:conflict_TTC}

% \STATE Extract bipartite graph $G = (X \cup B, E)$ with ILP-derived features $H_X, H_B, H_E$ as defined in \cref{BB}

% \STATE \textbf{// GNN Prediction}
% \STATE Feed $G$ into trained GNN model to obtain per-variable probabilities $\{p_i^k\}$

% % \FOR{each via $v_i$}
% %     \STATE $p_i \leftarrow \sum_{k=1}^{K} p_i^k$
% % \ENDFOR

% \STATE \textbf{// Temperature-Scaled Sampling}
% \FOR{each via $v_i$}
% \STATE Compute sampling probability $\tilde{p}_i$ using Eq.~\eqref{eq:temp_sampling}
% \ENDFOR

% \STATE Sample $\lfloor \frac{2}{3}|\mathcal{J}| \rfloor$ via indices from $\mathcal{J}$ without replacement according to $\{\tilde{p}_i\}$ 
% % \COMMENT{$\mathcal{J}$: via index set of the current layout}

% \STATE Let $\mathcal{R}^l \leftarrow$ indices of \textit{unsampled} (fixed) vias

% \STATE \textbf{// Constrained Re-optimization}
% \STATE Add fixing constraints to ILP using Eq.~\eqref{eq9}:
% \STATE \quad $\sum_{j:\, v_i \in t_j} t_j^{k_i} + v_i^{k_i} = 1, \quad \forall i \in \mathcal{R}^l$

% \STATE Solve reduced ILP (Sec.~\ref{sec:graph_model}) over remaining free vias

% \RETURN Final solution $\{v_i^k\}$
% \end{algorithmic}
% \end{algorithm}

\subsection{Inference Procedure}

\begin{table*}[!t]
\setlength{\tabcolsep}{7pt}
\centering
\caption{Comparison results of DSA with TPL on 7~nm synthetic layouts.}
\resizebox{0.92\textwidth}{!}{
\begin{tabular}{c|c c|c c c c|c c c c|c c c c} 
\hline
\multirow{2}{*}{Benchmark} 
 & \multicolumn{2}{c|}{Original} & \multicolumn{4}{c|}{EILP\cite{lin2019novel}} & \multicolumn{4}{c|}{DLXplus\cite{lin2019novel}} & \multicolumn{4}{c}{Ours} \\
\cline{2-15}
 & via & edge & cn & tn & cost & time (s) & cn & tn & cost & time (s) & cn & tn & cost & time (s) \\
\hline
test1  & 3977 & 7140 & 103 & 137 & 104.37 &  268.63 & 261 & 128 & 262.28 & 104.44 & 103 & 145 & 104.45 & 105.94 \\
test2  & 3807 & 6570 &  91 & 117 &  92.17 & 6655.68 & 220 &  98 & 220.98 &  85.18 &  91 & 120 &  92.20 &  50.09 \\
test3  & 3795 & 6170 &  81 &  99 &  81.99 &  446.88 & 195 &  79 & 195.79 &  80.31 &  81 & 102 &  82.02 &  31.90 \\
test4  &12025 &19882 & 265 & 314 & 268.14 &11528.07 & 542 & 114 & 243.14 & 612.33 & 265 & 324 & 268.24 & 193.88 \\
test5  & 5180 & 8587 & 127 & 110 & 128.10 &   76.65 & 265 & 112 & 266.12 & 189.71 & 127 & 113 & 128.13 &  84.02 \\
test6  & 6236 &10149 & 129 & 143 & 130.43 &  333.97 & 282 & 105 & 283.05 & 189.50 & 129 & 147 & 130.47 &  86.40 \\
test7  & 6303 &10472 & 151 & 158 & 152.58 &12658.32 & 325 & 196 & 326.96 & 173.73 & 151 & 163 & 152.63 &  60.13 \\
test8  & 5761 & 9402 & 114 & 135 & 115.35 &  141.16 & 339 & 159 & 340.59 & 152.93 & 114 & 146 & 115.46 &  44.99 \\
test9  & 4048 & 6699 &  98 &  94 &  98.94 & 1166.54 & 198 & 120 & 199.20 &  57.23 &  98 &  96 &  98.96 &  41.38 \\
test10 & 5157 & 8441 & 104 & 123 & 105.23 &  160.51 & 243 & 149 & 244.49 &  99.16 & 105 & 131 & 106.31 &  39.57 \\
\hline
Average &     &      &     &     & 127.73 & 3343.64 &     &     & 258.26 & 174.45 &     &     & 127.89 &  73.83 \\
Ratio   &     &      &     &     &  1.000 &   1.000 &     &     &  2.022 &  0.052 &     &     &  1.001 &  0.022 \\
\hline
\end{tabular}}
\label{main results}
\end{table*}

\begin{table*}[!t]
\setlength{\tabcolsep}{7pt}
\centering
\caption{Comparison results of DSA with TPL on 7~nm BOOM core layouts}
\resizebox{0.92\textwidth}{!}{
\begin{tabular}{c|c c|c c c c|c c c c|c c c c} 
\hline
\multirow{2}{*}{Benchmark} 
 & \multicolumn{2}{c|}{Original} & \multicolumn{4}{c|}{EILP\cite{lin2019novel}} & \multicolumn{4}{c|}{DLXplus\cite{lin2019novel}} & \multicolumn{4}{c}{Ours} \\
\cline{2-15}
 & via & edge & cn & tn & cost & time (s) & cn & tn & cost & time (s) & cn & tn & cost & time (s) \\
\hline
BOOM\_test1  & 15608 & 24758 & 273 & 210 & 275.10 &  634.57 &  900 &  774 &  907.74 &  706.67 & 273 & 439 & 277.39 & 121.29 \\
BOOM\_test2 & 16729 & 26325 & 430 & 209 & 432.09 & 1551.12 & 1133 & 1003 & 1143.03 & 1012.96 & 432 & 598 & 437.98 &  96.59 \\
BOOM\_test3 & 11586 & 18492 & 266 & 116 & 267.16 &  305.98 &  747 &  522 &  752.22 &  107.73 & 268 & 360 & 271.60 &  58.88 \\
BOOM\_test4 & 31486 & 50113 & 918 & 377 & 921.77 & 4920.61 & 2177 & 1899 & 2195.99 & 2051.60 & 922 & 938 & 931.38 & 330.97 \\
BOOM\_test5 & 13540 & 21320 & 285 & 156 & 286.56 & 1176.25 &  881 &  782 &  888.82 &  538.01 & 285 & 427 & 289.27 &  76.72 \\
BOOM\_test6 & 29442 & 46670 & 875 & 344 & 878.44 & 2235.45 & 1997 & 1719 & 2014.19 & 1665.30 & 877 & 781 & 884.81 & 324.96 \\
BOOM\_test7 & 16504 & 25903 & 424 & 198 & 425.98 & 1027.32 & 1061 &  940 & 1070.40 &  906.98 & 424 & 494 & 428.94 & 117.41 \\
BOOM\_test8 & 19691 & 30939 & 349 & 269 & 351.69 & 1121.64 & 1241 & 1124 & 1252.24 &  494.09 & 350 & 586 & 355.86 & 141.22 \\
BOOM\_test9 & 29672 & 46414 & 743 & 410 & 747.10 & 9929.41 & 2065 & 1741 & 2082.41 & 1780.18 & 746 & 885 & 754.85 & 296.56 \\
BOOM\_test10 & 23495 & 37210 & 665 & 290 & 667.90 & 2244.34 & 1660 & 1431 & 1674.31 &  896.68 & 666 & 844 & 674.44 & 154.30 \\
\hline
Average      &       &       &     &     & 525.38 & 2514.67 &      &      & 1398.14 & 1016.02 &     &     & 530.65 & 171.89 \\
Ratio        &       &       &     &     &  1.000 &   1.000 &      &      &   2.661 &   0.404 &     &     &  1.010 &  0.068 \\
\hline
\end{tabular}}
\label{main results BOOM}
\end{table*}

During inference, the trained GNN assigns to each via variable $v_i^k$ a probability $p_i^k$ of requiring reassignment. A hard threshold is typically  applied to reassign only those variables whose probability exceeds it. However, a hard threshold cannot adapt well to diverse layouts. Thus, we adopt a sampling-based strategy to adaptively select reassignment candidates.
Specifically, for each via $v_i$, we aggregate probabilities over masks and apply temperature scaling to obtain its sampling probability:
\begin{equation}
\label{eq:temp_sampling}
\tilde{p}_i = \frac{(p_i + \varepsilon)^{1/\tau}}{\sum_{j \in \mathcal{J}} (p_j + \varepsilon)^{1/\tau}},
\end{equation}
% where $p_i = \sum_{k=1}^{K} p_i^k$, 
% $\mathcal{J}$ denotes the set of all via indices in the layout library, and $\tau$ controls the sharpness of the probability distribution.
where $p_i = \sum_{k=1}^{K} p_i^k$, 
$\mathcal{J}$ denotes the set of all via indices in the current input layout, and $\tau$ controls the sharpness of the probability distribution.
We use a small $\tau=0.1$ to obtain a sharper distribution, making high-confidence vias more likely to be selected. According to $\tilde{p}_i$, we sample two-thirds of all vias from the layout without replacement and identify the remaining vias in each layout as fixed. 
For each layout $l$, the index set of fixed vias is denoted by $\mathcal{R}^l$. The ILP formulation in \Cref{sec:graph_model} is then extended by adding the following constraint to fix the variables associated with vias in $\mathcal{R}^l$:
\begin{equation}
\label{eq9}
\sum_{j: v_i \in t_j} t_j^{k_i} + v_i^{k_i} = 1, 
\quad \forall i \in \mathcal{R}^l,
\end{equation}
where $k_i$ denotes the mask index assigned to via $v_i$ in the initial solution. 
This effectively reduces the search space of the ILP solver, thereby accelerating its convergence. 
% The overall inference procedure is summarized in Algorithm~\ref{alg:inference}.

% \pgfplotsset{compat=1.18}
\section{Experimental Results}
\label{exp}
\subsection{Experimental Setting}

% We implemented our algorithm in Python, using Gurobi as the ILP solver~\cite{optimization2020gurobi}, and conducted all experiments on a machine equipped with an Intel Xeon Gold 5220 CPU @ 2.20 GHz. The number of available masks $K$=3. The $d_{\min}$ is set to 50~nm. The feasible templates include 1-hole and 2-hole 1-D straight templates. The weights in the ILP objective are set to $w_c = 1$ and $w_t = 0.01$, as in \cite{lin2019novel}. 

% For confidentiality, only a 7~nm layout clip from the industrial logic SoC design is available, containing approximately 3,000 vias. To reduce overfitting, we follow \cite{yang2019automatic} to generate 20 synthetic layouts track by track using design rules from the industrial layout, such as wire width $w=21$~nm and track pitch $p=31.5$~nm, for training. The GNN model was trained on a single NVIDIA A100 GPU with the Adam optimizer (learning rate $1\times10^{-3}$) in 46~s. For testing, 15 additional layouts of varying sizes were generated and combined with the industrial layout to evaluate accuracy and generalization. The full test datasets and design rules are available at \href{https://anonymous.4open.science/r/7nm-layouts-for-DSA_MP-3F7C}{\textit{https://anonymous.4open.science/r/7nm-layouts-for-DSA\_MP-3F7C}}

We implement our algorithm in Python, using Gurobi as the ILP solver~\cite{optimization2020gurobi}, and conduct all experiments on a machine equipped with an Intel Xeon Gold 5220 CPU @ 2.20~GHz. The number of available masks is set to $K=3$ for the 7~nm node and $K=4$ for the 5~nm node. The minimum edge-to-edge template spacing is $d_{\min} = 50$~nm. The feasible templates include 1-hole and 2-hole 1-D straight templates. The weights in the ILP objective are set to $w_c = 1$ and $w_t = 0.01$, following~\cite{lin2019novel}.

The GNN model comprises 2 interleaved half-convolutions~\cite{gasse2019exact} with all embeddings projected to dimension 64, followed by a 2-layer MLP with sigmoid activation. It is trained on a single NVIDIA A100 GPU using the Adam optimizer with a learning rate of $1\times10^{-3}$ for 200 epochs, completing in approximately 46~s. Details of the training and test sets are provided in Section~\ref{sec:benchmark}.

\subsection{Benchmark Details}
\label{sec:benchmark}

\textbf{Training Set.} Following~\cite{yang2019automatic}, we generate 20 synthetic layouts at the 7~nm node using design rules extracted from the open-source ASAP7 process design kit (PDK). Each layout contains approximately 3,000 vias, yielding a training set of roughly 60,000 vias in total. 
% The layouts are randomly split into 16 for training and 4 for validation (80\%/20\%). 
Label generation follows the procedure in \Cref{BB}, taking approximately 1 hour.

\textbf{Test Set.} To evaluate the generalizability of our method across different designs within the same technology node as well as across different technology nodes, we construct a diverse test set as follows.
\begin{itemize}
\item \textbf{7~nm Node (Synthetic):} 12 synthetic layouts generated following~\cite{yang2019automatic} using design rules from the ASAP7 PDK.
\item \textbf{7~nm Node (Open-source Processor Design):} 12 layouts extracted from a RISC-V BOOM core design \cite{celio2017boomv2} generated by Chipyard~\cite{amid2020chipyard} using the ASAP7 PDK, reflecting realistic design patterns from an open-source processor.
\item \textbf{5~nm Node (Industrial):} 11 clips from industrial logic SoC designs to evaluate cross-node generalization.
\end{itemize}
The full test datasets and layout generation scripts are available at \url{https://github.com/fengyu-neu/DSA-MP-Benchmark}.

\begin{table*}[!t]
\setlength{\tabcolsep}{7pt}
\centering
\caption{Comparison results of DSA with QPL on 5~nm industrial layouts.}
\resizebox{0.92\textwidth}{!}{
\begin{tabular}{c|c c|c c c c|c c c c|c c c c} 
\hline
\multirow{2}{*}{Benchmark} 
 & \multicolumn{2}{c|}{Original} & \multicolumn{4}{c|}{EILP\cite{lin2019novel}} & \multicolumn{4}{c|}{DLXplus\cite{lin2019novel}} & \multicolumn{4}{c}{Ours} \\
\cline{2-15}
 & via & edge & cn & tn & cost & time (s) & cn & tn & cost & time (s) & cn & tn & cost & time (s) \\
\hline
N5\_test1  & 4328  &  8244 &  27 & 288 &  29.88 &  299.25 & 115 & 255 & 117.55 &  50.26 &  27 & 294 &  29.94 &  44.51 \\
N5\_test2  & 7829  & 14772 &  40 & 540 &  45.40 & 2035.96 & 181 & 515 & 186.15 & 409.49 &  41 & 545 &  46.45 &  55.33 \\
N5\_test3  & 7006  & 12840 &  40 & 442 &  44.42 & 1230.44 & 151 & 435 & 155.35 & 196.12 &  40 & 453 &  44.53 &  58.89 \\
N5\_test4  & 10268 & 19226 &  55 & 741 &  62.41 &  541.88 & 247 & 703 & 254.03 & 549.43 &  55 & 749 &  62.49 &  88.20 \\
N5\_test5  & 6671  & 12262 &  25 & 436 &  29.36 &  653.43 & 134 & 408 & 138.08 & 147.88 &  25 & 441 &  29.41 &  48.95 \\
N5\_test6  & 9986  & 18539 &  39 & 684 &  45.84 & 1690.79 & 216 & 622 & 222.22 & 474.95 &  39 & 694 &  45.94 & 164.20 \\
N5\_test7  & 10432 & 19543 &  61 & 691 &  67.91 & 2246.64 & 242 & 654 & 248.54 & 565.55 &  62 & 703 &  69.03 & 142.48 \\
N5\_test8  & 8082  & 15044 &  39 & 509 &  44.09 &  600.20 & 174 & 480 & 178.80 & 249.38 &  41 & 515 &  46.15 & 276.02 \\
N5\_test9  & 7284  & 13316 &  27 & 442 &  31.42 & 2158.09 & 142 & 418 & 146.18 & 181.39 &  28 & 447 &  32.47 &  52.36 \\
N5\_test10 & 7637  & 13373 &  34 & 410 &  38.10 &  227.93 & 145 & 414 & 149.14 & 175.29 &  37 & 437 &  41.37 &  99.91 \\
\hline
Average    &       &       &     &     &  43.883 & 1168.460 &     &     & 179.604 & 299.973 &     &     &  44.778 & 103.085 \\
Ratio      &       &       &     &     &   1.000 &    1.000 &     &     &   4.093 &   0.257 &     &     &   1.020 &  0.088 \\
\hline
\end{tabular}}
\label{N5 results}
\end{table*}

\begin{table*}[!t]
\setlength{\tabcolsep}{7pt}
\centering
\caption{Cases where some algorithms exceed the 20-hour time limit (``TLE'').}
\resizebox{0.92\textwidth}{!}{
\begin{tabular}{c|c c|c c c c|c c c c|c c c c} 
\hline
\multirow{2}{*}{Benchmark} 
 & \multicolumn{2}{c|}{Original} 
 & \multicolumn{4}{c|}{EILP\cite{lin2019novel}} 
 & \multicolumn{4}{c|}{DLXplus\cite{lin2019novel}} 
 & \multicolumn{4}{c}{Ours} \\
\cline{2-15}
 & via & edge & cn & tn & cost & time (s) & cn & tn & cost & time (s) & cn & tn & cost & time (s) \\
\hline
N5\_test\_11  & 5551   & 11120   & TLE & TLE & TLE & TLE & 161   & 403   & 165.03   & 122.05    & 34    & 526   & 39.26    & 57.51     \\
test11        & 116215 & 206000  & TLE & TLE & TLE & TLE & TLE   & TLE   & TLE      & TLE       & 3622  & 3895  & 3660.95  & 4636.34   \\
test12        & 206659 & 366628  & TLE & TLE & TLE & TLE & TLE   & TLE   & TLE      & TLE       & 6287  & 6421  & 6351.21  & 17265.78  \\
BOOM\_test11  & 221696 & 352602  & TLE & TLE & TLE & TLE & TLE   & TLE   & TLE      & TLE       & 6054  & 5572  & 6109.72  & 5856.67   \\
BOOM\_test12  & 667477 & 1056475 & TLE & TLE & TLE & TLE & TLE   & TLE   & TLE      & TLE       & 18315 & 18787 & 18502.87 & 49369.75  \\
\hline
\end{tabular}}
\renewcommand{\arraystretch}{1.0}
\label{TLE}
\vspace{-2mm}
\end{table*}
\subsection{Result Analysis}
We evaluate our method against state-of-the-art decomposition approaches on two configurations: DSA with triple patterning lithography (DSA-TPL) at the 7~nm node, and DSA with quadruple patterning lithography (DSA-QPL) at the 5~nm node. In all tables, ``via'' denotes the number of vias, ``edge'' the number of spacing conflict edges, ``cn'' the number of unresolved conflicts, ``tn'' the number of two-hole templates, ``cost'' the total cost defined in \Cref{eq:obj}, and ``time'' the runtime in seconds. The reported time for our method includes initial ILP solving, GNN inference, and final re-optimization. 

We compare against the EILP method proposed in~\cite{lin2019novel}, which guarantees optimal cost but is often prohibitively slow, and its heuristic acceleration ``DLXplus,'' which improves runtime at the expense of solution quality. The results and analysis are as follows.

\textbf{DSA-TPL (7~nm).} To evaluate generalization across different design styles within the same technology node, we test on 7~nm layouts from two sources as described in \Cref{sec:benchmark}: synthetic layouts and layouts extracted from a RISC-V BOOM core design. Results are reported in \Cref{main results} and \Cref{main results BOOM}, respectively. In these benchmarks, the ratio of conflict edges to vias is close to 2:1, significantly higher than the 1:2 ratio reported in prior work~\cite{lin2019novel}, reflecting the denser conflict relationships typical of advanced-node designs.

As shown in \Cref{main results}, on synthetic layouts our method achieves about $45\times$ speedup over EILP with only 0.1\% increase in cost. On the BOOM layouts (\Cref{main results BOOM}), our method still provides about $15\times$ acceleration with around 1.0\% cost increase. These results demonstrate that our approach maintains substantial runtime advantages while preserving near-optimal solution quality across different design styles at the 7 nm node.

\textbf{DSA-QPL (5~nm).} To evaluate cross-node generalization, we test our method on 5~nm industrial layouts using DSA-QPL. The results are summarized in \Cref{N5 results}. The GNN model employed shares identical weights with that used in \Cref{main results,main results BOOM}, without any retraining or fine-tuning on 5~nm data. Despite the domain shift in both technology node and number of masks, our method still achieves approximately $12\times$ speedup over EILP with only 2\% increase in cost, further validating the cross-node transferability of the learned GNN model.

\textbf{Performance on Challenging Layouts.} \Cref{TLE} reports cases across all test sets where existing methods exceed the 20-hour time limit. Our method remains tractable on these challenging instances, whereas EILP or DLXplus exceed the time limit without producing feasible solutions. Even for the instance where DLXplus produces a feasible solution (N5\_test\_11), our method still achieves a $4.2\times$ lower cost within less than half the runtime. The experimental results demonstrate the robustness and effectiveness of our proposed method on layouts that are challenging due to either large scale or complex conflict structures.

% \textbf{Performance on Challenging Layouts.} \Cref{tab:tle} reports cases where existing methods exceed the 20-hour time limit. Our method handles all instances within the limit, while EILP or DLXplus fail to produce feasible solutions. This demonstrates the robustness of our local refinement strategy on layouts that are challenging due to either large scale or complex conflict structures.

% \textbf{Generalization.} Our GNN model is trained on 20 synthetic layouts of approximately 3,000 vias each at the 7~nm node. Despite this limited training data, it generalizes well across multiple dimensions: (1)~to larger layouts with up to \_\_\_ vias, (2)~to different design sources including open-source processor designs and industrial layouts, and (3)~to a different technology node (5~nm) with a different mask count ($K=4$). These results indicate that the structural patterns learned by the GNN transfer effectively across layout scales, design styles, and technology configurations.
\textbf{Generalization Analysis.}
Our GNN model is trained on 20 synthetic layouts containing about 60,000 vias at the 7~nm node. The model generalizes well across multiple dimensions: (1)~to larger layouts, e.g., layouts with over 200K vias,  (2)~to unseen layouts from diverse origins, including the BOOM core and industrial designs, (3)~to a different technology node (5~nm), and (4)~to a different mask count ($K=4$).

This strong generalization mainly stems from our model operating on ILP representations rather than layout-specific geometric features. The input features are derived from the ILP formulation and the solver, including objective coefficients, constraint biases, constraint coefficients, and solution-pool frequencies. These features reflect how variables influence the objective and constraints during optimization, providing solver-level information that is largely independent of the specific layout topology or design style.

In addition, the bipartite graph representation of the ILP together with message passing allows the GNN to operate on graphs of varying sizes, enabling the model to transfer to larger layouts. Finally, our framework adopts a GNN-guided local refinement strategy, where the GNN predicts vias that are likely to require reassignment, while the remaining variables are fixed to their initial assignments. The ILP solver then determines the final mask assignments by optimizing only over the selected subset of variables. Since the actual mask assignment is still decided by the ILP solver, the solution quality is preserved while the search space is significantly reduced, which further contributes to the robustness of our framework.

\subsection{Ablation Study}

\begin{figure}[t]
\centering
\begin{tikzpicture}
\begin{axis}[
    ybar,
    bar width=0.25cm,
    width=0.52\columnwidth,
    height=4.5cm,
    ymin=0.92, ymax=1.00,
    ytick={0.92,0.94,0.96,0.98,1.00},
    yticklabel style={font=\small},
    symbolic x coords={w/o coef, w/o sol\_pool\_freq, w/o point\_num, Full},
    xtick=data,
    xticklabel style={font=\small, rotate=30, anchor=east},
    ylabel={Recall},
    ylabel style={font=\small},
    ymajorgrids=true,
    grid style=dashed,
    enlarge x limits=0.15,
    axis y line*=left,
    axis x line*=bottom,
]
\addplot[fill=blue!40] coordinates {
    (w/o coef,             0.944)
    (w/o sol\_pool\_freq,  0.933)
    (w/o point\_num,       0.962)
    (Full,                 0.979)
};
\end{axis}
\end{tikzpicture}
\hfill
\begin{tikzpicture}
\begin{axis}[
    ybar,
    bar width=0.25cm,
    width=0.48\columnwidth,
    height=4.5cm,
    ymin=127.0, ymax=129.5,
    ytick={127.0,127.5,128.0,128.5,129.0,129.5},
    yticklabel style={font=\small},
    symbolic x coords={w/o coef, w/o sol\_pool\_freq, w/o point\_num, Full},
    xtick=data,
    xticklabel style={font=\small, rotate=30, anchor=east},
    ylabel={Cost},
    ylabel style={font=\small},
    ymajorgrids=true,
    grid style=dashed,
    enlarge x limits=0.15,
    axis y line*=left,
    axis x line*=bottom,
]
\addplot[fill=orange!40] coordinates {
    (w/o coef,            128.83)
    (w/o sol\_pool\_freq, 129.05)
    (w/o point\_num,      128.77)
    (Full,                127.89)
};
\end{axis}
\end{tikzpicture}
\vspace{-0.3cm}
\caption{Ablation study on solver-derived features.}
% \vspace{-0.3cm}
\label{fig:ablation}
\end{figure}
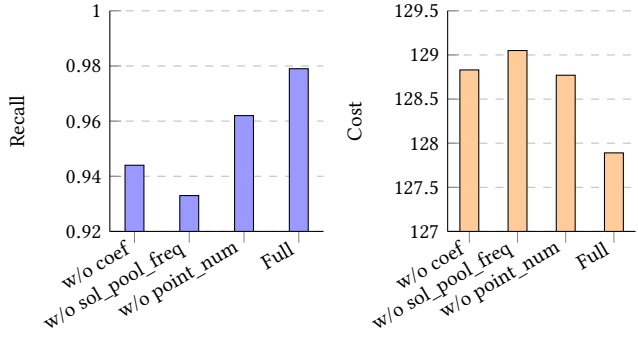

\textbf{Solver-Derived Features.} To assess the impact of each solver-derived feature in $H_X$, we conduct an ablation study by removing each feature in turn and evaluating on all layouts listed in \Cref{main results}, with all metrics averaged across the test layouts. The results are shown in \Cref{fig:ablation}. We report \textit{recall} and \textit{cost} as the primary metrics: recall measures the fraction of vias requiring reassignment that are correctly identified by the GNN model. High recall is essential to avoid missing vias that require reassignment, which is critical for achieving optimal solutions. Cost reflects the final solution quality after our local refinement strategy. 

Removing any single feature degrades both cost and recall, confirming that all features contribute to model performance. Excluding \texttt{sol\_pool\_freq} results in the largest degradation in both metrics, reducing recall from $0.979$ to $0.933$ and increasing cost from $127.89$ to $129.05$, indicating that solution-pool frequency is the most informative feature for capturing solver behavior. Excluding \texttt{coef} and \texttt{point\_num} results in smaller but consistent degradation across both metrics. Overall, the full feature set achieves the best performance.

% \begin{table}[!t]
% \label{data sources ablation}
% \centering
% \caption{GNN performance on 5~nm industrial layouts with different training sets.}
% % \vspace{-2mm}
% \label{tab:dataset_ablation}
% \begin{tabular}{lcccc}
% \toprule
% Training Data & cn & dsa & cost & time (s) \\
% \midrule
% the BOOM core layouts & 44.4 & 556.7 & 49.967 & \textbf{97.218} \\
% Synthetic layouts      & \textbf{39.5} & \textbf{527.8} & \textbf{44.778} & 103.085 \\
% \bottomrule
% \end{tabular}
% \end{table}

\renewcommand{\arraystretch}{1.0}

\begin{table}[!t]
\centering
\caption{GNN performance on 5~nm industrial layouts with different training sets.}
% \vspace{-2mm}
\label{data sources ab}
\begin{tabular}{lcccc}
\toprule
Training Data & cn & tn & cost & time (s) \\
\midrule
the BOOM core layouts & 44.4 & 556.7 & 49.967 & \textbf{97.218} \\
Synthetic layouts      & \textbf{39.5} & \textbf{527.8} & \textbf{44.778} & 103.085 \\
\bottomrule
\end{tabular}
\end{table}
\textbf{Training Set.} To study the impact of training data source, we construct an alternative training set from the BOOM core layouts described in \Cref{sec:benchmark}, extracting a comparable scale of 60,000 vias. Apart from the training data source, all model architectures and training settings remain identical. Both models are evaluated on the 5~nm industrial benchmark under the DSA-QPL configuration, and results are summarized in \Cref{data sources ab}. 

The model trained on synthetic layouts achieves lower cost than the one trained on the BOOM core layouts. This advantage likely stems from the diversity introduced during synthetic generation: randomized wire lengths, spacings, and via placements under design-rule constraints produce conflict graphs with diverse local structures, which in turn yield ILP formulations with varied constraint interaction patterns. Such diversity helps the model learn solver-level behaviors that generalize better to unseen layouts. Furthermore, synthetic layouts can be generated efficiently without relying on proprietary design data, making our approach practical for settings where real-world layouts are unavailable.

\begin{figure}[!t]
\centering
\begin{tikzpicture}
\begin{axis}[
    ybar,
    bar width=0.28cm,
    width=0.48\columnwidth,
    height=4.5cm,
    ymin=220, ymax=275,
    ytick={220,230,240,250,260,270},
    yticklabel style={font=\small},
    symbolic x coords={No DSA, Prior Model, Ours},
    xtick=data,
    xticklabel style={font=\small, rotate=30, anchor=east},
    ylabel={EPE},
    ylabel style={font=\small},
    ymajorgrids=true,
    grid style=dashed,
    enlarge x limits=0.25,
    axis y line*=left,
    axis x line*=bottom,
]
\addplot[fill=blue!40] coordinates {
    (No DSA,269.3)
    (Prior Model,227.2)
    (Ours,225.3)
};
\end{axis}
\end{tikzpicture}
\hfill
\begin{tikzpicture}
\begin{axis}[
    ybar,
    bar width=0.28cm,
    width=0.48\columnwidth,
    height=4.5cm,
    ymin=4.08, ymax=4.17,
    ytick={4.08,4.10,4.12,4.14,4.16},
    yticklabel style={font=\small},
    symbolic x coords={No DSA, Prior Model, Ours},
    xtick=data,
    xticklabel style={font=\small, rotate=30, anchor=east},
    ylabel={$L_2$ Loss ($\times10^6$ nm$^2$)},
    ylabel style={font=\small},
    ymajorgrids=true,
    grid style=dashed,
    enlarge x limits=0.25,
    axis y line*=left,
    axis x line*=bottom,
]
\addplot[fill=orange!40] coordinates {
    (No DSA,4.159)
    (Prior Model,4.096)
    (Ours,4.091)
};
\end{axis}
\end{tikzpicture}
\vspace{-0.3cm}
\caption{Manufacturability comparison among no DSA (MPL only), prior conflict graph model, and our proposed conflict graph model.}
\label{fig:manufacturability}
\end{figure}
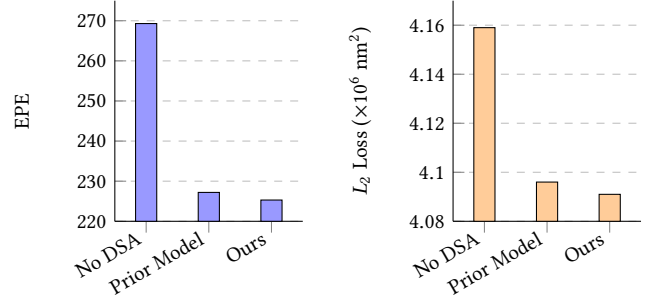

\begin{figure}[!t]
\centering
\includegraphics[width=0.95\linewidth]{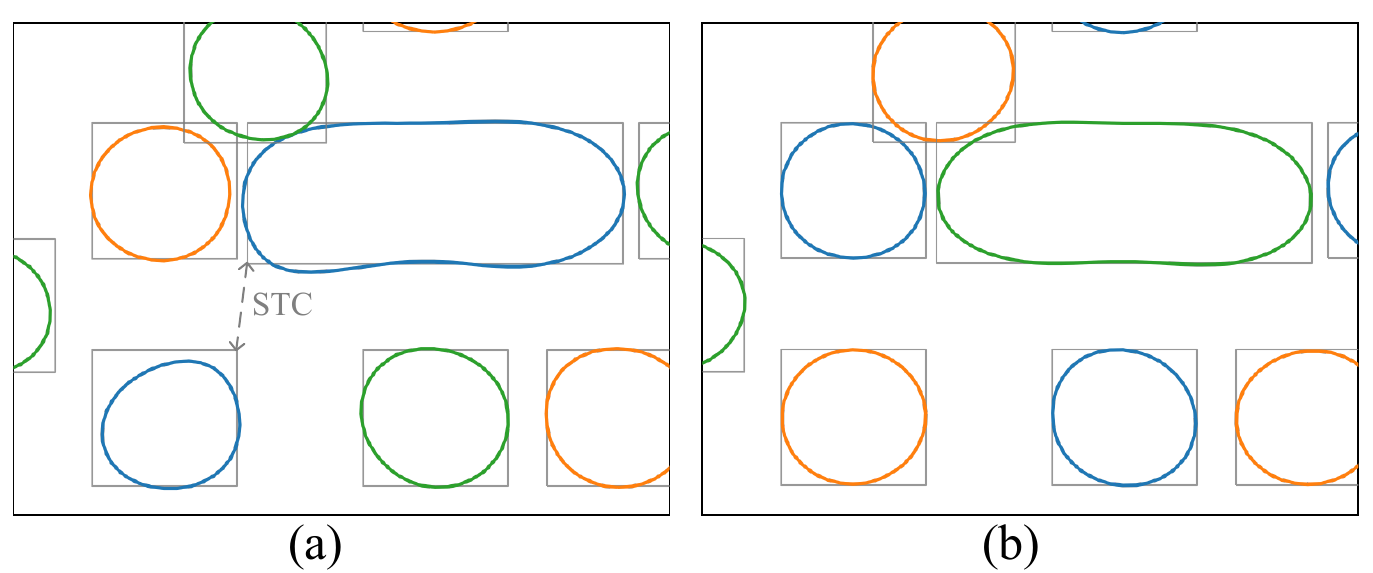}
\caption{Post-OPC lithography simulation results. Gray rectangles denote targets and colored contours represent simulated printed patterns on different masks. (a) Results obtained using the prior conflict graph model. (b) Results obtained using our proposed conflict graph model.} 
\vspace{-5mm}
\label{post-opc}
\end{figure}
\textbf{Conflict Graph Model.} We evaluate manufacturability under three settings shown in \Cref{fig:manufacturability}. For each setting, we perform OPC and lithography simulation using a commercial tool, measuring average Edge Placement Error (EPE)~\cite{banerjee2013iccad} and $L_2$ loss across layouts in \Cref{main results}. EPE counts the number of sampled edge points where the deviation between target and printed pattern exceeds a threshold (5~nm in our experiments). $L_2$ loss measures the squared difference between the printed and target images. Our proposed conflict graph model (based on edge-to-edge template spacing) achieves the lowest EPE and $L_2$ loss among all settings. \Cref{post-opc} further shows post-OPC simulation results. The proposed conflict graph model produces more accurate contours for DSA templates, enabling higher-quality via formation through directed self-assembly.

\section{Conclusion}
\label{con}
We propose a new conflict graph model that explicitly accounts for edge-to-edge spacing constraints between DSA templates, which were neglected in prior work. We also present a GNN-guided local refinement framework for DSA-MP layout decomposition, achieving significant speedup over state-of-the-art methods with high-quality solutions. 
Our framework is general and has the potential to be applied to other ILP-based optimization problems in physical design.

% \begin{acks}
% \end{acks}

\bibliographystyle{ACM-Reference-Format}
% \bibliography{MPL/ref/sample-base}
\bibliography{MPL/ref/Top,MPL/ref/main}

\appendix

\end{document}